# Learnable Heterogeneous Convolution: Learning both topology and strength

Rongzhen Zhao, Zhenzhi Wu *, Qikun Zhang

*Lynxi Technologies, Beijing 100097, China*

**ABSTRACT**

Existing convolution techniques in artificial neural networks suffer from huge computation complexity, while the biological neural network works in a much more powerful yet efficient way. Inspired by the biological plasticity of dendritic topology and synaptic strength, our method, Learnable Heterogeneous Convolution, realizes joint learning of kernel shape and weights, which unifies existing handcrafted convolution techniques in a data-driven way. A model based on our method can converge with structural sparse weights and then be accelerated by devices of high parallelism. In the experiments, our method either reduces VGG16/19 and ResNet34/50 computation by nearly 5× on CIFAR10 and 2× on ImageNet without harming the performance, where the weights are compressed by 10× and 4× respectively; or improves the accuracy by up to 1.0% on CIFAR10 and 0.5% on ImageNet with slightly higher efficiency. The code will be available on www.github.com/Genera1Z/LearnableHeterogeneousConvolution.

## 1. Introduction

Convolution neural networks (CNNs) are showing their superior performance in vision tasks like classification, detection and segmentation, but their advantages in practice are impeded by their heavy computation.

To improve CNN efficiency, researchers have developed many recipes: (1) compressing existing models with pruning (Liu et al., 2019; Zhou, Zhang, Wang, & Tian, 2019), quantization (Cao et al., 2019; Gong et al., 2019) or knowledge distillation (Jin et al., 2019; Peng et al., 2019); (2) designing efficient network structures either by hand (Hu, Shen, & Sun, 2018) and Ma, Zhang, Zheng, and Sun (2018) or by automatic search (Chen, Xie, Wu, & Tian, 2019; Howard et al., 2019); (3) exploiting efficient convolution operators, which are either smaller (Simonyan & Zisserman, 2014), or factorized (Szegedy, Ioffe, Vanhoucke, & Alemi, 2016; Szegedy, Vanhoucke, Ioffe, Shlens, & Wojna, 2016), or sparse (Huang, Liu, M.L.V., & Weinberger, 2018; Sun, Li, Liu, & Wang, 2018). On the other hand, (4) the neural network in a brain, which features sparse activation and high dynamics (Holtmaat et al., 2005; Stettler, Yamahachi, Li, Denk, & Gilbert, 2006), works in a much more complex and powerful yet efficient way (Merolla et al., 2014).

For recipe (1), a large model of high performance must be available firstly, and the pruning-retrain iteration is very time-consuming; for recipe (2), it is hard to design an excellent structure by hand, and also too costly to search one even with tens of GPUs and days. So we choose to combine recipes (3) and (4), namely, sparsifying convolutions by learning from the brain.

A biological neuron connects from multiple predecessor neuron via dendrites, and to a subsequent neuron via an axon. The connections are plastic both in topology and strength, through forming/losing/developing the synapses on dendritic spines (Bhatt, Zhang, & Gan, 2009; Harms & Dunaevsky, 2007). To make an analogy (Beysolow II, 2017), like Fig. 1, for the $c_o$ kernels in a standard convolution layer, each kernel of shape $(k, k, c_i)$ belongs to a neuron, which is shared at different spatial positions of feature maps; for the $c_i$ slices in a kernel, each kernel slice of shape $(k, k)$ is a dendrite; for the $k * k$ elements in a slice, each element is a synapse. But unlike biological neurons, in a standard convolution layer, only the strength of kernel elements, i.e., weights is learnable, while the topology of kernel slices is not, let alone the number of slices and kernels. Such differences are where the CNN can learn from the brain.

Improved convolution techniques like Group/Depth/Point-Wise Convolution (GWC, DWC, PWC) can be seen as sparsification of neurons in a layer or dendrites in a neuron, which are, however, predefined instead of learnt (Alex, Sutskever, & Hinton, 2012; Sandler, Howard, Zhu, Zhmoginov, & Chen, 2018); works like SeeSaw and DeepR introduce in learnability (Guillaume, David, Wolfgang, & Robert, 2018; Zhang, 2019). Other works like

* Corresponding author.
   *E-mail addresses:* rongzhen.zhao@lynxi.com (R. Zhao), zhenzhi.wu@lynxi.com (Z. Wu), qikun.zhang@lynxi.com (Q. Zhang).

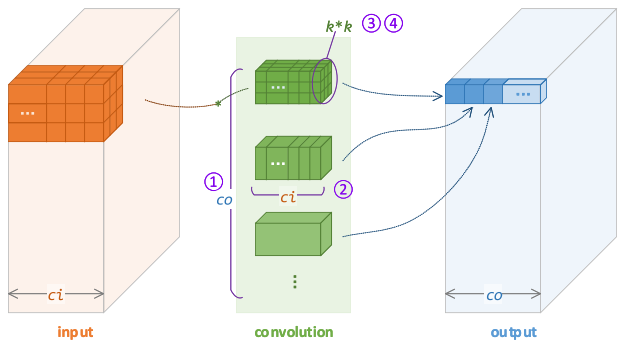

**Fig. 1.** A convolution layer (green, the biases and activation are ignored) between input feature maps (orange) and output features (blue). For the standard convolution, ① the number of kernels in a layer is fixed $c_o$, ② the number of slices in each kernel is fixed $c_i$, ③ the shape of each kernel slice (topology) is fixed $(k, k)$, and only ④ the value of the $k*k$ elements (strength) in each kernel slice can be learnt. Our method makes ①~④ all learnable. (For interpretation of the references to color in this figure legend, the reader is referred to the web version of this article.)

Inception and HetConv sparsify the dendritic topology square to a row, line or dot, which are also not learnable (Singh, Verma, Rai, & Namboodiri, 2019; Szegedy et al., 2015); works like L0 training hit the mark by a fluke, but the weights learnt is non-structural, which is not friendly for hardware to accelerate (Christos, Max, & Diederik, 2018). See Section 2 for detailed reviewing.

Our method, Learnable Heterogeneous Convolution (LHC), a seamless replacement to the standard convolution, is proposed to break those limitations by integrating the plasticity in both the synaptic strength, i.e., learnable weights, which is originally possessed by the standard convolution, and additionally the dendritic topology, i.e., learnable kernel slice shapes. With the latter as a unit, the number of slices in a neuron, and the number of neurons in a layer all become learnable.

Comprehensive experiments demonstrate the superiority of our method. At parallelism = 512, suppose batch size is 1, we can either reduce VGG/ResNet computation by nearly 5× on CIFAR10 and 2× on ImageNet without harming the performance, where the weights are compressed by 10× and 4× respectively; or improve the accuracy by up to 1.0% on CIFAR10 and 0.5% on ImageNet with slightly higher efficiency. The extra costs are no more than slightly longer training time.

Our contributions are:

(1) LHC is proposed to realize dual-plasticity of strength and topology in convolution, which is fine-grainedly yet structurally sparse thus fits hardware acceleration;

(2) It sparsifies convolutions in full back-propagation, instead of undifferentiable ways used by most pruning methods such as cutting of weights that have least magnitude;

(3) It requires negligible extra costs at training and can greatly improve CNN models' efficiency at inference stage even with some performance gain.

The remaining content is organized as follows: related works are reviewed in Section 2; the proposal is detailed in Section 3; how our method unifies various convolution techniques is analyzed in Section 4; experiments are presented in Section 5; more advanced analyses are discussed in Section 6; the conclusion is drawn in Section 7.

Given a convolution layer, the notations list in Table 1.

**Table 1**
Notations used in this article.

| | |
|---|---|
| Shape of the kernels | $(k, k, c_i, c_o)$ |
| Shape of input feature maps or "feat in" | $(b, h_i, w_i, c_i)$ |
| Shape of output feature maps or "feat out" | $(b, h_o, w_o, c_o)$ |
| Batch dimension of feature maps | $b$ |
| Number of input/output channels | $c_i, c_o$ |
| Height/width of input/output features | $h_i, w_i, h_o, w_o$ |
| Spatial size of a kernel | $k*k, (k, k)$ |
| Topology constraint of input/output channels | $c_{gi}, c_{go}$ |
| Tera Operations Per Second | TOPS |
| Floating Point Operation(s) | FLOP(s) |
| Multiply Accumulate (unit) | MAC |

## 2. Related works

Here the sparse convolution techniques for improving CNN efficiency are reviewed, and keypoints where the convolution can imitate the brain are inducted.

**Structural Sparse Convolutions: Handcrafted**

Convolution can be sparsified in spatial dimensions. A 2D convolution kernel is factorized into two perpendicular 1D ones in Szegedy, Ioffe, Vanhoucke, and Alemi (2016) and Szegedy, Vanhoucke, Ioffe, Shlens, and Wojna (2016). Kernels are designed into incremental sizes in Tan and Le (2019). Slices of a 3*3 kernel are replaced by 1*1 sizes at intervals in Singh et al. (2019) shown in Fig. 8.

Sparsification can also be taken in channels. In a GWC (SIfre & Mallat, 2014), the input channels are grouped and each output channel is correlated with one of the groups. Features among different groups are further exchanged in Sun et al. (2018), Xie et al. (2018) and Zhang, Qi, Xiao, and Wang (2017). Shifting or negation saves kernels/channels either, like in Shang, Sohn, Almeida, and Lee (2016) and Yan, Li, Li, Zuo, and Shan (2018).

Dimensions of space and channel can be considered together. Like in Wang, Xu, Chunjing, Xu, and Tao (2018), standard kernels are turned into secondary kernels, which are degressive in space and partially connected in channel.

Neuron number in a layer, dendrite number in a neuron, and dendritic topology are all handcrafted in such methods; only the synaptic strength is learnable.

**Structural Sparse Convolutions: Learnt**

Compared with the above, works thinking in this way bring in learnability of sparsity. But the techniques employed are still restricted to GWC, DWC and PWC.

Works like Huang et al. (2018) require too many iterations, where connections between input and output channels of less importance are cut off progressively while training to get a desired group number. Some realize learnability under the guidance of Singular Value Decomposition like Peng et al. (2018), which is built of GWC and PWC in a way similar to Howard et al. (2017). Others like Zhang (2019) realize unevenly grouped GWC layers in a model by fusing network architecture search into training.

Only sparsity in channel dimension is considered by them. Namely, besides the synaptic strength, they do take into account the plasticity of neurons number in a layer or dendrite number in a neuron, but overlook the dendritic topology.

By the way, Verelst and Tuytelaars (2020) is worth learning from, which generates feature masks to skip computations corresponding to zeros, even if it sparsifies activations instead of weights.

**Non-Structural Sparse Convolutions: Learnt**

Such works are similar to the pruning methods (Liu et al., 2019; Zhou et al., 2019), except that the sparsity is gained during rather than after training. Since non-structural sparsity goes against hardware acceleration (Deng, Li, Han, Shi, & Xie, 2020), recent works are not that many.

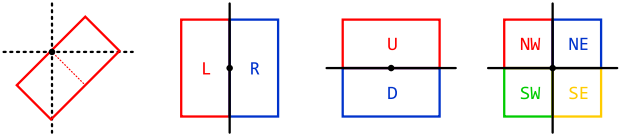

**Fig. 2.** Kernels of uncommon shapes designed handcrafted by SWF (Yin et al., 2019) to address specific image contents, which inspired our work.

L0 regularization are often used to train models full of zeros, like Christos et al. (2018), enabling joint optimization of weights' value and topology via non-negative random gates of hard concrete distribution. Inspired by dynamic connections in the brain Bhatt et al. (2009) and Harms and Dunaevsky (2007), the rewiring mechanism is proposed by Guillaume et al. (2018) to enable simultaneous learning of connections and weights for constrained hardware resources.

Their implementation of plasticity in dendritic topology, namely, neuron number in a layer or dendrite number in a neuron, are good references.

Our inspiration: making it possible in a convolution layer to (a) learn the dendritic topology and synaptic strength at the same time, and (b) realize fine-grained yet structural sparsity for hardware acceleration.

## 3. Proposed method

How LHC integrates the plasticity in dendritic topology and synaptic strength is elaborated here. The structural sparsification and hardware acceleration of LHC-based models are also described.

### 3.1. Prototype

**Convolution Kernels of Uncommon Shapes**

For either traditional or CNN-based algorithms, it is quite common to use convolution kernels of square shapes. However, SWF (Yin, Gong, & Qiu, 2019), a latest work, broke fresh ground and is equal in force compared with CNN-based algorithms. The key is that kernels of uncommon shapes, shown in Fig. 2, were meticulously designed for different image patterns to gain excellent feature extraction capability.

Enlightened by this, we empirically design 15 rigid kernel shapes, shown in Fig. 3(a). Among them, shape ⟨1⟩1, ⟨2⟩1 and ⟨6⟩1 are common in use; shape ⟨3⟩1 and ⟨3⟩3 are already used in GoogLeNet Inceptions; shape group ⟨4⟩ and ⟨5⟩ inherit from SWF. Shape group ⟨2⟩∼⟨5⟩ are designed to sparsify a convolution layer in spatial dimensions, while group ⟨1⟩ is to reduce redundancy in channel dimension.

The theoretical foundation for such a design is that CNNs are able to extract visual patterns of different levels like dot, edge, curve and plane at different layers (Goodfellow, Bengio, & Courville, 2016). Accordingly, very sparse shape groups ⟨2⟩ and ⟨3⟩ are designed to address simple patterns like dots and edges; shape groups ⟨4⟩∼⟨6⟩, not that sparse, are to handle complex patterns like curves and planes.

Further on, we relieve the constraints of the aforementioned rigid shapes, and for a 3*3 kernel, we provide $2^{3*3} = 512$ free shapes for an LHC layer to choose, shown in Fig. 3(b). Obviously, the rigid shapes are a sub-set of the free shapes.

An LHC layer equipped with the rigid shapes is called LHCR, and LHC with the free shapes is LHCF. In LHC, these shapes are formed as masks consisting of 0/1 elements, and are paired with corresponding effect factors to guide the learning of the shapes of kernel slices. Details are presented in Section 3.3.

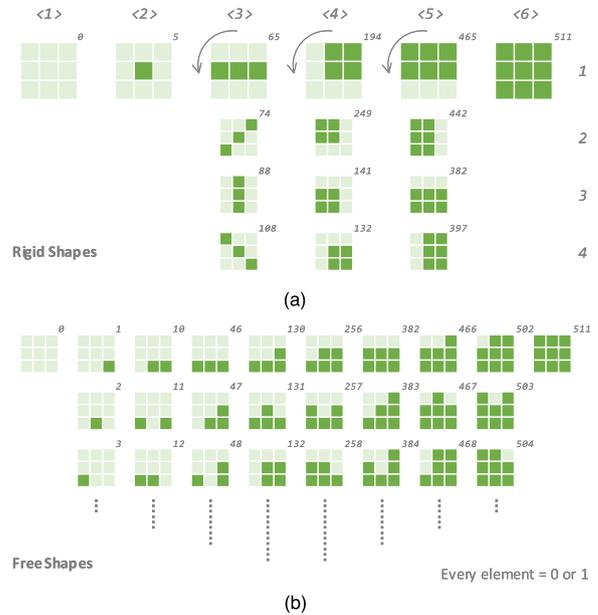

**Fig. 3.** Two sets of uncommon shapes for LHC. (a) 15 rigid shapes: designed empirically in six groups; digits on the top and right are their indexes, e.g., ⟨3⟩4. (b) $2^{3\times 3} = 512$ free shapes: generated by arbitrarily setting each element to 0 or 1; digits on every top-right corner are their serial indexes, e.g., 502. The former is a sub-set of the latter.

To make a comparison, LHCR implies priori knowledge of those rigid shapes while LHCF has more freedom in sparsification, which is verified by experiments in Section 5.

**Learnable Heterogeneous Convolution**

For a Learnable Heterogeneous Convolution (LHC) layer, its kernel slices are learnt to be these uncommon shapes, rather than pre-defined.

Suppose a convolution layer with $c_i$ input channels and $c_o$ output channels. With the aforementioned uncommon shapes, the kernels can be shaped (1) kernel-by-kernel (KbK), where shapes are the same within a kernel and different among kernels, or (2) slice-by-slice (SbS), where shapes are different both within a kernel and among kernels. Refer to Fig. 4 the top two rows.

Obviously, SbS is more flexible than KbK and thus can eliminate the convolution redundancy in both spatial and channel dimensions better; but the computation graph of SbS is too fragmented for the hardware to accelerate. So we split the difference: only one shape combination is allowed for every adjacent $c_{gi}$ slices and every adjacent $c_{go}$ kernels. Then we can learn a structural sparse LHC layer, which supports parallelism up to $c_{gi}*c_{go} \equiv 64*8$. Refer to the bottom two rows of Fig. 4. Besides, these constraints happen to be a kind of regularization, benefiting the performance. See Section 6 for details.

### 3.2. Computation reduction

To facilitate discussion, we take the 3*3 convolution as an example, and use the indexes in Fig. 3 to represent shapes.

Following the aforementioned setting, the computation quantity or the total number of multiplication–addition of a standard convolution layer is:

$$\begin{aligned} C_{\text{STD}} &= h_o \times w_o \times c_i \times c_o \times \|s^{511}\|_0 \\ &= h_o \times w_o \times c_i \times c_o \times 9 \end{aligned} \quad (1)$$

where $s^{511}$ the 511th shape in Fig. 3 and its L0 norm is 9.

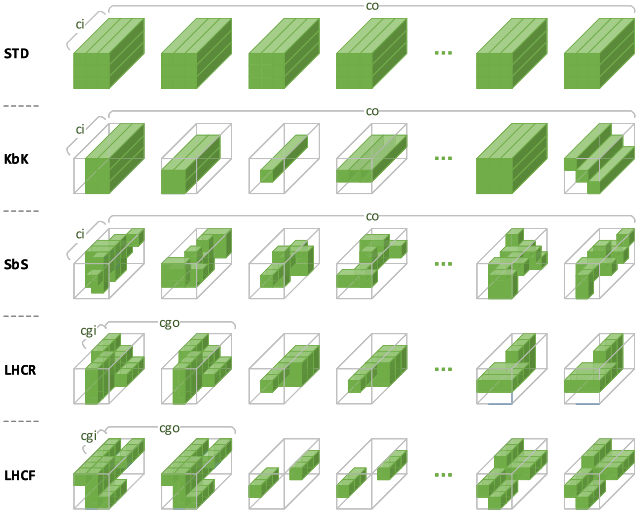

**Fig. 4.** Topology of LHC Kernels. 1st row: LHC-KbK, where kernel slices have the same shapes within a kernel and different shapes among kernels; 2nd row: LHC-SbS, where slices have different shapes both within and among kernels; 3rd row: LHCR with topology constraints $c_{gi}$ and $c_{go}$, where slices are shaped into those rigid shapes and every $c_{gi} * c_{go}$ slices have the same shape; 4th row: LHCF, where slices are shaped into those free shapes.

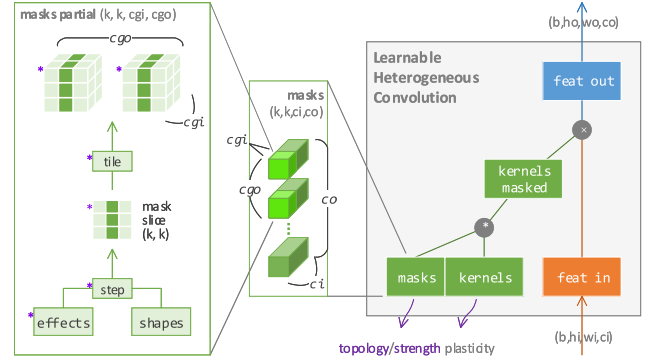

**Fig. 5.** Training principle of LHC. Left box: (1) every $c_{gi} * c_{go}$ kernel slices are equipped with a set of effect factors, each of which points to a shape belonging to the rigid shapes or the free; (2) through a differentiable step function, a mask slice is got then tiled into masks partial of shape $(k, k, c_{gi}, c_{go})$. Center box: do (2) for every $c_{gi}*c_{go}$ slices in the kernels, then the masks of shape $(k, k, c_i, c_o)$ are constructed. Right box: (3) multiply the masks with the kernels element-wisely; (4) do the convolution with the masked kernels and the input feature maps of shape $(b, h_i, w_i, c_i)$, finally get the output features of shape $(b, h_o, w_o, c_o)$. (5) In the backward propagation, the shapes get different gradients according to their contributions. (6) As training goes on, the advantage of different shapes is accumulated in the effect factors, and the most suitable shapes gradually win out. Note: in the left box, purple stars are where gradients pass during back propagation; purple stars of the other two boxes are omitted; in the right box, the biases and activation are omitted for simplicity.

Similarly, the computation quantity of an LHC layer is:

$$C_{\text{LHC}} = h_o \times w_o \times \left( \sum_{y=1}^{c_o/c_{go}} \left( \sum_{x=1}^{c_i/c_{gi}} \|s_{x,y}\|_0 \times c_{gi} \right) \times c_{go} \right) \\ = h_o \times w_o \times c_{gi} \times c_{go} \times \sum_{s=1}^{c_i/c_{gi} \times c_o/c_{go}} n_s \quad (2)$$

where $c_{gi}$ and $c_{go}$ are the aforementioned topology constraints; $s_{x,y}$ is the shape of the $x$th slice in the $y$th kernel; $n_s$ is L0 norm of the $s$th shape in all $c_i/c_{gi} \times c_o/c_{go}$ positions, i.e., $\|s_{x,y}\|_0 = n_s$.

So, the computation reduction is:

$$\Delta C = h_o \times w_o \times (c_i \times c_o \times 9 \\ - c_{gi} \times c_{go} \times \sum_{s=1}^{c_i/c_{gi} \times c_o/c_{go}} n_s) \quad (3)$$

when $n_s = 9$, $\Delta C = 0$, where all shapes in LHC is $s^{511}$, i.e., the standard convolution; when $n_s = 0$, $\Delta C = C_{\text{STD}}$, where all shapes in LHC is $s^0$, which means all features are redundant and thus can only be approximated.

Besides, existing convolution techniques can be unified by LHC. See Section 4 for details.

### 3.3. Learnability

At the training stage, the only difference between LHC and a standard convolution is the construction of the kernels. As shown in Fig. 5, the masks composed of zeros and ones are constructed first, then multiplied with the kernels, where elements multiplied with zeros in the masks are deactivated during both forward and back-propagation; finally, the masked kernels are convolved with the input features, and the output features are obtained. The masks and kernels realize plasticity in dendritic topology and synaptic strength respectively.

**Guide the Learning of Topology**

For a convolution layer, its computation quantity is proportional to the density of its kernels; for a CNN model, its computation quantity is positively correlated with its global density, of which the maximum is 1, namely, no sparsification, and the minimum limit is 0.

Given a model with $L$ LHC layers, we can set a global density target and calculate mask regularization loss; then by minimizing it, we can guide the model to converge to a state that costs much less computation:

$$l_{\text{mask}} = |d_t - \frac{\sum_{l=1}^{L} \|M_l\|_1}{\sum_{l=1}^{L} \text{size}(M_l)}| \quad (4)$$

where $l_{\text{mask}}$ is the mask regularization loss; $d_t$ is the density target; $M_l$ is the masks in the $l$th LHC layer.

Here, L1 norm is used instead of L0, because (1) $M_l$ only consists of zeros and ones, which means L0 is equal to L1; and (2) L1 norm is differentiable.

Hence, optimizing this model turns into simultaneously minimizing the mask regularization loss and the task loss, e.g., categorical cross-entropy loss:

$$\min_{M_1,\ldots,M_L} \alpha \times l_{\text{mask}} + l_{\text{task}} \quad (5)$$

$$\text{s.t.} \quad c_{gi} = C_1, c_{go} = C_2 \quad (6)$$

where $\alpha$ is a positive constant; $l_{\text{task}}$ is the task loss; $c_{gi}$ and $c_{go}$ are the aforementioned topology constraints for structural sparsity, and their value $C_1 \equiv 64$ and $C_2 \equiv 8$ typically.

Now the point is how to construct the $M_l$.

Given the kernels of a layer and the topology constraints $c_{gi}$ and $c_{go}$, to construct the masks, we have $c_i/c_{gi} \times c_o/c_{go}$ positions to determine. In other words, for each of these positions, we need to calculate a mask slice, a $k*k$ matrix, using the aforementioned shapes that are either rigid or free.

**Enroll Shapes into Competition**

To calculate a mask slice, we hope those shapes illustrated in Fig. 3 are enrolled in all together so that they compete with one another along with the training steps, letting the fittest eventually win out.

For **LHCR**, at each of these positions, a 15D vector $e$ called effect factors is used to represent the effects of those 15 rigid

shapes. So the mask slice for this position is:

$$\boldsymbol{m} = \sum_{i=1}^{15} \text{step}(e_i, \boldsymbol{e}) \times s_i \quad (7)$$

$$\text{step}(e_i, \boldsymbol{e}) = \begin{cases} 1 & e_i = \max(\boldsymbol{e}) \\ 0 & e_i < \max(\boldsymbol{e}) \end{cases} \quad (8)$$

$$\nabla \text{step}(e_i, \boldsymbol{e}) \equiv \begin{cases} 1 & |e_i - \text{mean}(\boldsymbol{e})| < 1 \\ c & \text{others} \end{cases} \quad (9)$$

where $e_i$ is the $i$th element in $\boldsymbol{e}$ and $s_i$ is the $i$th of those 15 rigid shapes; $c$ is a small positive constant, empirically set to 0.1; step is differentiable and ensures the summation of effect factors be one. softmax is not used because it is heavy to calculate and can hardly make one of the shapes win out.

For **LHCF**, equip each of these positions with a 3*3 matrix $\boldsymbol{e}$ as the effect factors, which represent the effects of those 512 free shapes for this position. So the mask slice for this position is:

$$\boldsymbol{m} = \text{step}(\boldsymbol{e}) \quad (10)$$

$$\text{step}(e_{i,j}) = \begin{cases} 1 & e_{i,j} > 0 \\ 0 & e_{i,j} \leq 0 \end{cases} \quad (11)$$

$$\nabla \text{step}(e_{i,j}) \equiv \begin{cases} 1 & |e_{i,j}| < 1 \\ c & \text{others} \end{cases} \quad (12)$$

where $i, j$ indexes the elements in matrix $\boldsymbol{e}$; $c$ is a small positive constant, empirically set to 0.1.

Both of the $\nabla$step functions are designed to have grad = 1 and grad = 0.1 segments, so that under Xavier-like initialization, which is widely adopted, the shape of each mask slice is actively learnable at the early stage of training and becomes more stable but not absolutely stable at the late stage of training. Refer to Section 6 for more details.

Via the aforementioned formulas, a mask slice $\boldsymbol{m}$ is calculated, then tiled/repeated in the dimensions of input and output channel $c_{gi}$ and $c_{go}$ times respectively, then get the masks partial $M_{\text{part}}$ of shape $(3, 3, c_{gi}, c_{go})$ for this position. By iterating all those $c_i/c_{gi} \times c_o/c_{go}$ positions, the masks $M$ of shape $(3, 3, c_i, c_o)$ are finally constructed.

**Smoothing the Training Process**

To smooth the training process, two warm-up tricks are employed.

---

**Algorithm 1** Mask Enabling Warm-Up

Set the number of epochs $n_{\text{warm}}$ for warm-up
1. calculate the incremental unit per epoch $\delta = 1/n_{\text{warm}}$
2. on the $i$th epoch begins: enable every mask at probability $p_i = \delta * (i - 1)$
3. conduct this training epoch
4. turn to step 2 until $n_{\text{warm}}$ reached

---

**Algorithm 2** Mask Regularizing Warm-Up

Set the number of epochs $n_{\text{warm}}$, the weight for mask regularization $\alpha_t$, target density $d_t$
1. calculate the incremental unit per epoch $\delta = \alpha_t/n_{\text{warm}}$
2. conduct a training epoch, and keep the task loss $l_{\text{task}}$
3. calculate the upper limit and the scale factor of this loss: $l_{\max} = \text{abs}(1.0 - d_t)$, $f = l_{\text{task}}/l_{\max}$
4. on epoch $i$ begins: update the weight $\alpha = f * \delta * (i - 1)$
5. conduct this training epoch
6. turn to step 4 until $n_{\text{warm}}$ is reached

---

Algorithm 1: mask enabling warm-up, enables the masks in LHC layers randomly with increasing possibilities at the beginning epochs, to eliminate non-convergence issues.

Algorithm 2: mask regularizing warm-up, increases the weight of $r_{mask}$ gradually at the starting epochs, to let the CNN model learn strength first and then topology.

**Extra Cost Analysis**

Compared with the standard convolution, the extra weights, i.e., the effect factors, brought in by an LHC layer during training is $3 * 3 * c_i * c_o/(c_{gi} * c_{go})/(3 * 3 * c_i * c_o) = 1/(c_{gi} * c_{go})$. Suppose $c_{gi} = 64$ and $c_{go} = 8$, 0.1953% of extra storage is paid for the topology plasticity.

The extra computation, i.e., the construction of the masks, brought in by an LHC layer during training is $3 * 3 * c_i * c_o/(h_o * w_o * 3 * 3 * c_i * c_o) = 1/(h_o * w_o)$. Suppose a 224*224 input image and a VGG16 model, the computation increases by about 0.3856%.

So at the training stage, the cost of our method is nearly the same as that of standard convolution. See Section 5 for more details.

### 3.4. Acceleration

At the inference stage, CNN models built of LHCs can be easily accelerated by hardware of high parallelism, which consumes much less memory and computation.

Given a MAC array of parallelism = 512, i.e., 512 multiplication and addition units. To maximize hardware utilization, two groups of 512 numbers must be fed to them in one clock period, which means such numbers must be stored continuously respectively. This can be achieved under "structural sparsity", but cannot in fragmented computation graphs. This is the key point in implementing the optimal hardware acceleration.

**Efficient Hardware Implementation**

As mentioned above, the calculation of convolution is the only part influenced by LHC, so the following discussion focuses on this part.

To realize high parallel computing of a standard convolution layer, sufficient units of multipliers and adders to form a MAC array are needed. Similarly, to make the most of the structural sparsity of LHC layers, it is needed that

(1) In the weight buffer, only valid weights of LHC are loaded from external storage before runtime, and are stored in an intensive way. The memory consumption of an LHC layer is just about 50%~1% of that of a standard convolution layer, so bandwidth and computation at runtime are saved.

(2) In the I/O buffers, input and output feature maps are stored at runtime, just like that of standard convolution.

**Data Management and Parallel Computing**

Here explains how all invalid weights and invalid computation are avoided and how high parallelism is achieved.

As shown in Fig. 6, as well as Fig. 7, the input/output features are organized in a $c_{gi}$-aligned /$c_{go}$-aligned manner, i.e., the width of the IO buffers is $c_{gi}/c_{go}$; yet the weights are organized in a $c_{gi} * c_{go}$-aligned manner.

Given input features of shape $(h_i, w_i, c_i)$, for every pixel of shape $(c_i, )$ in it, every $c_{gi}$ adjacent elements are stored as a row in the input buffer. At each convolution step, elements of shape $(3, 3, c_i)$ in current sliding window are copied to the window buffer and are stored in the same way.

Given weights of shape $(3, 3, c_i, c_o)$, for each kernel slice group of shape $(3, 3, c_{gi}, c_{go})$ in it, every $c_{gi} * c_{go}$ elements are stored as a row in the weight buffer, with full-zero $c_{gi} * c_{go}$-length segments being skipped. At every convolution step, feature elements in the buffer window that correspond to these full-zero segments are skipped according to the discontinuous addresses provided by ALUT.

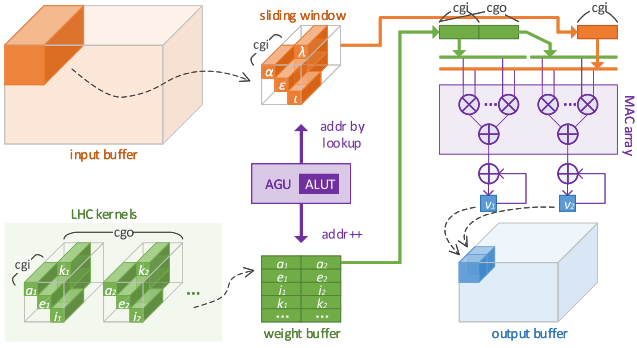

**Fig. 6.** Parallel processing of LHC inference on hardware — how all invalid weights and computation are avoided, and how high parallelism is reached. Input buffer: input features are stored in $c_{gi}$-aligned manner. Window buffer: elements in current sliding window are copied from the input buffer. Weight buffer: weights are stored in $c_{gi}*c_{go}$-aligned manner, with full-zero segments of length $c_{gi}*c_{go}$ being skipped. AGU: address generation unit, counts addresses for the weights buffer incrementally. ALUT: address look-up table, stores addresses calculated in advance for the window buffer. At every convolution step, every row of $c_{gi}*c_{go}$ weights in the weight buffer, and the corresponding row of length $c_{gi}$ in the window buffer, are sent to the MAC array; then the $c_{go}$ results are sent to $c_{go}$ output registers, where the value accumulates $3*3*c_i/c_{gi}$ times to get the final $c_{go}$ results in the output features. With topology constraints $c_{gi} \equiv 64$ and $c_{go} \equiv 8$, a 512-way parallel processing is achieved. If taking the batch dimension $b$ (not drawn for simplicity) into account, the parallelism can be further increased by $b$ times.

are fetched via the pre-defined address saved in ALUT (address lookup table).

With the addresses from AGU and ALUT, every row of length $c_{gi}*c_{go}$ in the weight buffer, and the corresponding row of length $c_{gi}$ in the window buffer, are sent to the MAC array to do parallel multiplication–addition; then the $c_{go}$ results are sent to $c_{go}$ output registers to accumulate $3*3*c_i/c_{gi}$ times; finally the output element at location $(x, y)$ in the corresponding channel of the output features is got. By iterating all the sliding windows and all the kernels, the full output features are got.

Under topology constraints $c_{gi}$ and $c_{go}$, for a kernel, every $c_{gi}$ adjacent kernel slices have the identical topology, and for a layer, every $c_{go}$ adjacent kernels have the identical topology. So at each clock cycle, $c_{gi}*c_{go}$ elements can be fed to the MAC array. Suppose $c_{gi}$ = 64 and $c_{go}$ = 8, parallelism = 512 is reached. This means there are 1024 multiplication and addition operations per clock, and the peak compute power reaches 1TOPS at 1 GHz clock frequency, which meets the compute requirement of most terminal devices specifically designed for CNN acceleration (Andri, Karunaratne, Cavigelli, & Benini, 2020; Moons, Bankman, Yang, Murmann, & Verhelst, 2018).

Of course the parallelism can be further improved if taking the batch dimension into account. Given batch size $b$, the parallelism can be further increased to $b*c_{gi}*c_{go}$.

**Extra Cost Analysis**

Compared with the standard convolution, extra components are required to accelerate an LHC layer during inference, i.e., ALUT for indexing sparse weights and AGU for skipping invalid feature elements.

Since employing different storing ways, the window buffer and input buffer need different fetch addresses. Data in the weights buffer are fetched via the incremental address counted by AGU (address generation unit); data in the window buffer

For ALUT, every $c_{gi}*c_{go}$ weights need an address index, and the overall density of all LHC layers is no greater than 20%, so extra memory consumption is just $k*k*c_i/c_{gi}*c_o/c_{go}/(k*k*c_i*c_o) = 1/c_{gi}/c_{go} = 0.1953\%$ of that of the standard convolution. Let alone 80+% of space is already saved by our method.

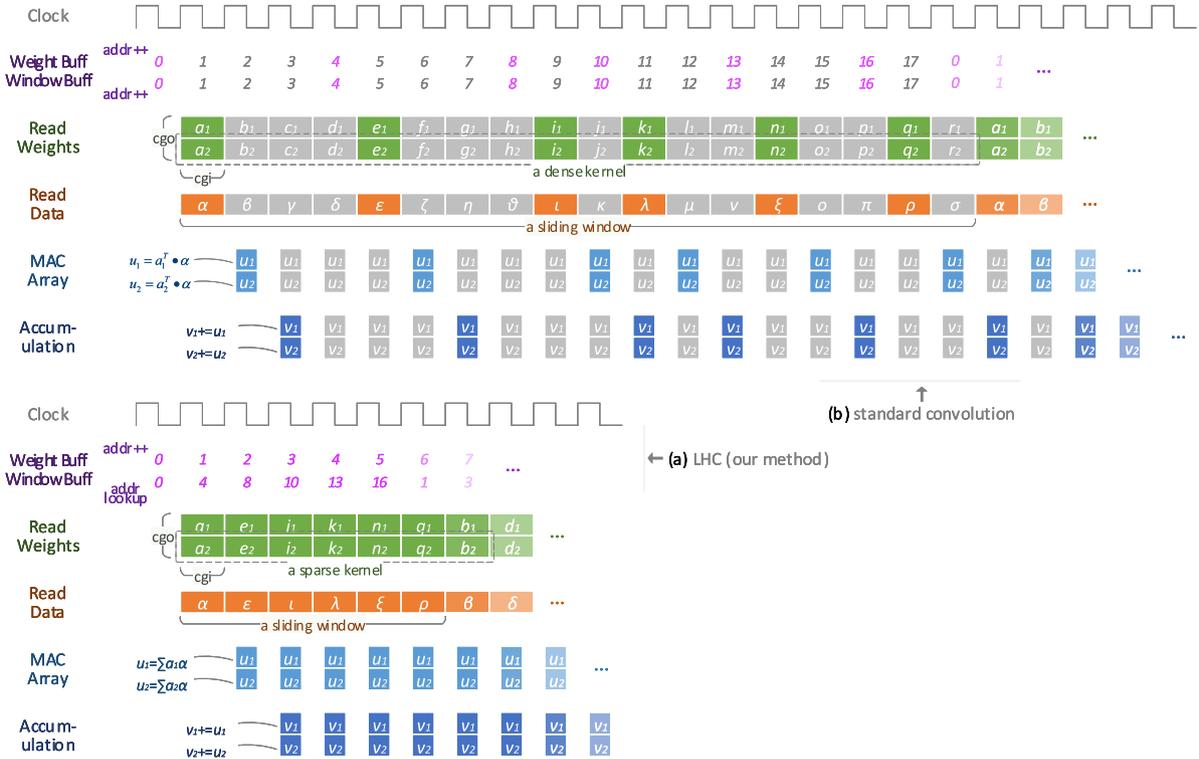

**Fig. 7.** Time scheduling of LHC vs traditional convolution. (a) The timing of an LHC layer, where its structural sparsity is fully utilized and thus all redundant clocks and other computational resources are saved. (b) The timing of a standard convolution layer, where sparsity (gray parts) is not considered. Each column is $c_{gi} \times c_{go}$ parallel operations such as read/write, multiply–add and accumulation. Refer to Fig. 6 for the corresponding hardware logic. (For interpretation of the references to color in this figure legend, the reader is referred to the web version of this article.)

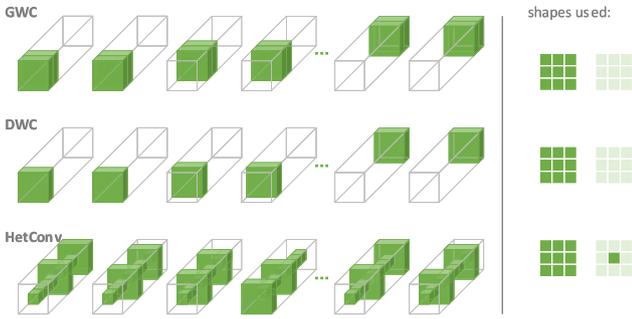

**Fig. 8.** The unification of various convolution techniques. GWC, DWC and HetConv can all be taken as LHCs that use some certain shapes.

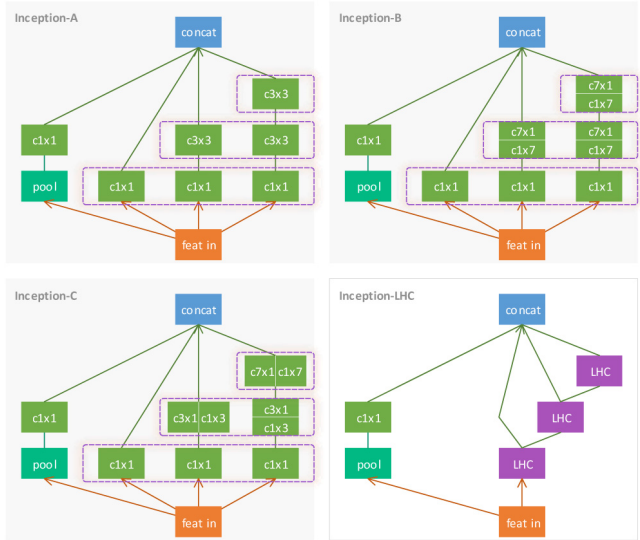

**Fig. 9.** Rebuilding GoogLeNet InceptionV4-A/B/C with LHC. The shapes used by LHCs are the ones used in the original Inceptions. (For interpretation of the references to color in this figure legend, the reader is referred to the web version of this article.)

For AGU, the implementation just requires negligible resources compared with the whole.

So at the inference stage, a large portion of computational resources can be saved, even though a little extra cost is introduced in.

## 4. Unification

With the topology plasticity, various convolution techniques discussed in Section 2 can be unified by LHC. As shown in Figs. 8 and 9, representative ones like GWC, DWC, HetConv and Inception are used as examples.

The shape of the $x$th slice in the $y$th kernel of a convolution layer is notated as $s_{x,y}$.

**Group-Wise Convolution**

Group-Wise Convolution or GWC (Alex et al., 2012), can be viewed as an LHC using shapes $\langle 1 \rangle 1$ and $\langle 6 \rangle 1$ only, illustrated in Fig. 8 the first row. Under the settings of Section 3.3, when the following conditions are satisfied, an LHC degenerates into a GWC:

$$s_{x,y} = \begin{cases} s^{511} & \left( \begin{array}{l} kc_{gi} < x \le (k+1)c_{gi} \\ kc_{go} < y \le (k+1)c_{go} \end{array} \right) \\ s^0 & \text{others} \end{cases} \quad (13)$$

where $c_{gi} = c_i/n_{\text{group}}$, $c_{go} = c_o/n_{\text{group}}$, $k = 0, 1, \ldots, n_{\text{group}}-1$.

**Depth-Wise Convolution**

Depth-Wise Convolution or DWC (Sandler et al., 2018), illustrated in Fig. 8 the second row, a special case of GWC, is an LHC using shapes $\langle 1 \rangle 1$ and $\langle 6 \rangle 1$ only. When the following conditions are satisfied, an LHC degenerates into a DWC:

$$s_{x,y} = \begin{cases} s^{511} & \left( \begin{array}{l} kc_{gi} < x \le (k+1)c_{gi} \\ kc_{go} < y \le (k+1)c_{go} \end{array} \right) \\ s^0 & \text{others} \end{cases} \quad (14)$$

where $c_{gi} = 1$, $c_{go} = c_o/c_i$, and $k = 0, 1, \ldots, c_i$-1.

**HetConv**

HetConv (Singh et al., 2019) is an LHC using shapes $\langle 2 \rangle 1$ and $\langle 6 \rangle 1$ only, illustrated in Fig. 8 3rd row. When the following conditions are satisfied, given $p$ as the number of shape $\langle 6 \rangle 1$ in a kernel, LHC degenerates into HetConv:

$$s_{x,y} = \begin{cases} s^{511} & (x+y-1)\%p = 0 \\ s^1 & \text{others} \end{cases} \quad (15)$$

where $c_{gi} = 1$, $c_{go} = 1$, and $p$ is a hyper parameter selected from $\{1, 2, \ldots, c_i\}$.

**Inception**

GoogLeNet Inceptions (Szegedy, Ioffe, Vanhoucke, & Alemi, 2016) can be rebuilt with LHC as shown in Fig. 9. Each dashed purple rectangle is equivalent to an LHC layer. For Inception-A, the 2nd and 3rd rectangles are GWCs where the groups are not evenly divided, and thus can be replaced by LHC. In Inception-B/C, orthogonal 1D convolution pairs are employed to approximate standard 2D convolutions, either serially or parallelly, where the former amounts to double non-linearity and cannot be equivalent to one LHC layer. But for simplicity, they are all replaced by LHC, then a unified Inception shown in the bottom right of Fig. 9 is got.

**Others**

Other convolution techniques designed for efficiency can also be unified. PWC can be viewed as standard convolution of kernel size 1*1; MixConv (Tan & Le, 2019) is a GWC variant where kernel sizes are different; SeeSaw (Zhang, 2019), IGC (Sun et al., 2018; Xie et al., 2018; Zhang et al., 2017), etc., are unevenly grouped GWCs.

## 5. Experiments

Here experiments are presented to compare LHC with other convolution techniques that are either widely used or achieve state-of-the-art results.

### 5.1. Experiment settings

Listed in Table 2 are the experiment items.

**Networks and Convolution Techniques**

The comparison of LHC, standard convolution, HetConv, FPGM (He et al., 2019) and Taylor (Molchanov et al., 2019) are conducted on VGG16/19 and ResNet34/50. Among them, HetConv represents techniques of structural sparse convolution; and FPGM/Taylor represent techniques of non-structural sparse convolution.

The comparison of LHC, Inception, GWC+channel-shuffle and DWC+PWC are conducted on GoogLeNet InceptionV4, ShuffleNetV1 and MobileNetV1 respectively.

**Switchover of Different Convolution Techniques**

For VGG16/19 and ResNet34/50, all convolution layers except the 1st are replaced by LHCR/LHCF and HetConv.

For GoogLeNet, Inceptions are rebuilt by LHCR in the way described in Fig. 9. Shapes used in LHCR are $\langle 1 \rangle 1$, $\langle 2 \rangle 1$, $\langle 6 \rangle 1$, $\langle 3 \rangle 1$ and $\langle 3 \rangle 3$.

For ShuffleNet, all GWC+channel-shuffle structures are replaced by LHCR, where shapes used are $\langle 1 \rangle 1$ and $\langle 6 \rangle 1$.

## Table 2
Experiment items: candidate convolution techniques on mainstream networks.

| | Origin | GWC (Alex et al., 2012) | DWC (Sandler et al., 2018) | Inception (Szegedy, Ioffe, Vanhoucke, & Alemi, 2016) | HetConv (Singh et al., 2019) | LHC (ours) | FPGM (He, Liu, Wang, Hu, & Yang, 2019) | Taylor (Molchanov, Mallya, Tyree, Frosio, & Kautz, 2019) |
|---|---|---|---|---|---|---|---|---|
| VGG16 (Simonyan & Zisserman, 2014) | ✓ | | | | ✓ | ✓ | ✓ | ✓ |
| VGG19 (Simonyan & Zisserman, 2014) | ✓ | | | | ✓ | ✓ | ✓ | ✓ |
| ResNet34 (He, Zhang, Ren, & Sun, 2016) | ✓ | | | | ✓ | ✓ | ✓ | ✓ |
| ResNet50 (He et al., 2016) | ✓ | | | | ✓ | ✓ | ✓ | ✓ |
| GoogLeNet (Szegedy, Ioffe, Vanhoucke, & Alemi, 2016) | – | | | ✓ | | ✓ | | |
| ShuffleNet (Zhang, Zhou, Lin, & Sun, 2018) | – | ✓ | | | | ✓ | | |
| MobileNet (Howard et al., 2017) | – | | ✓ | | | ✓ | | |

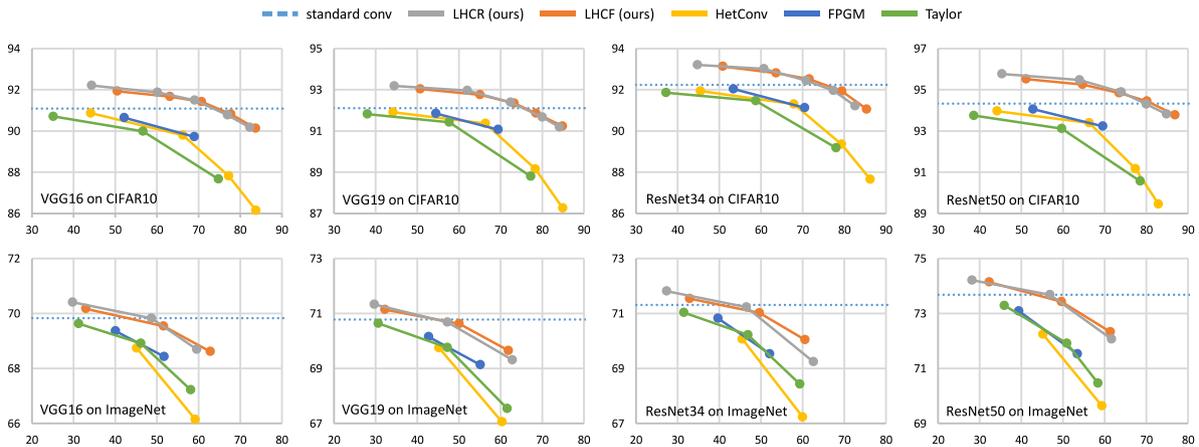

**Fig. 10.** Curves of computation reduction vs accuracy of different networks built of different techniques on different datasets. The x-axis $\Delta$flops is computation reduction in percentage, and the y-axis acc is classification accuracy in percentage. (For interpretation of the references to color in this figure legend, the reader is referred to the web version of this article.)

For MobileNet, all DWC+PWC structures are replaced by LHCR, where shapes used are $\langle 1 \rangle 1$, $\langle 2 \rangle 1$ and $\langle 6 \rangle 1$.

**Tasks for Making Evaluations**

The tasks for evaluating different convolution techniques are the image classification, on two widely recognized datasets, CIFAR10 and ImageNet.

Training procedures are designed to be identical: on CIFAR10/ImageNet, models are to be trained 200/100 epochs, with early stopping patience 40/20, using SGD with initial learning rate 1e−2/1e−3 and decay factor 0.1. Data augmentations include random flipping, translation, rotation, hue, saturation, brightness and contrast.

Specifically, for networks built of LHC,

(1) Hyper-parameter $d_t$, is set to {invalid, 0.2, 0.1, 0.05, 0.01} for CIFAR10, and {invalid, 0.25, 0.1} for ImageNet. Here invalid means no density target is set, and thus the network can explore the most suitable shapes and converge to the best performance.

(2) Hyper-parameter $c_{gi}$ and $c_{go}$ are set to 64 and 8 respectively as parallelism $\geq$ 512 is a typical value of hardware acceleration (Andri et al., 2020; Moons et al., 2018).

For FPGM and Taylor, experiments are carried out by following their official procedures.

### 5.2. Result analysis

Results are presented in computation reduction vs classification accuracy, as shown in Fig. 10.

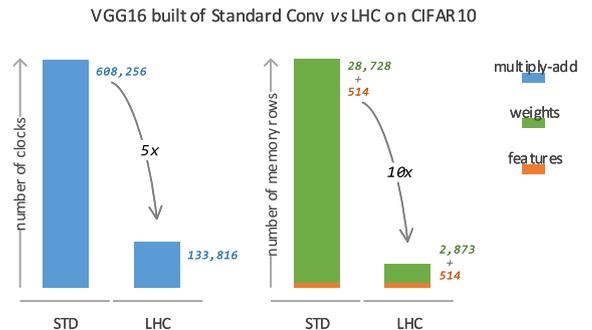

**Fig. 11.** Hardware resource saving with LHC. Clocks of 5× can be saved, which either benefits improving the throughput or reducing energy consumption. Memory of 10× can also be saved, which dramatically reduces the on-chip memory block resource. This model VGG16 is trained on CIFAR10 under constraints of parallelism = 512 and no obvious accuracy loss.

**LHC vs Standard Conv, HetConv, FPGM & Taylor**

Shown in Fig. 10 are the results of LHC and other convolution techniques. The x-axis is the computation reduction in percentage, and y-axis is the classification accuracy. The more a line to the top right corner, the better it is.

By comparing LHCR and LHCF, LHCR works slightly better if $\Delta$flops is smaller, while LHCF is a bit superior if $\Delta$flops is larger. The former can be attributed to the priori knowledge implied in the design of those rigid shapes, and the latter is likely due to the full freedom of those free shapes in dropping redundant weights.

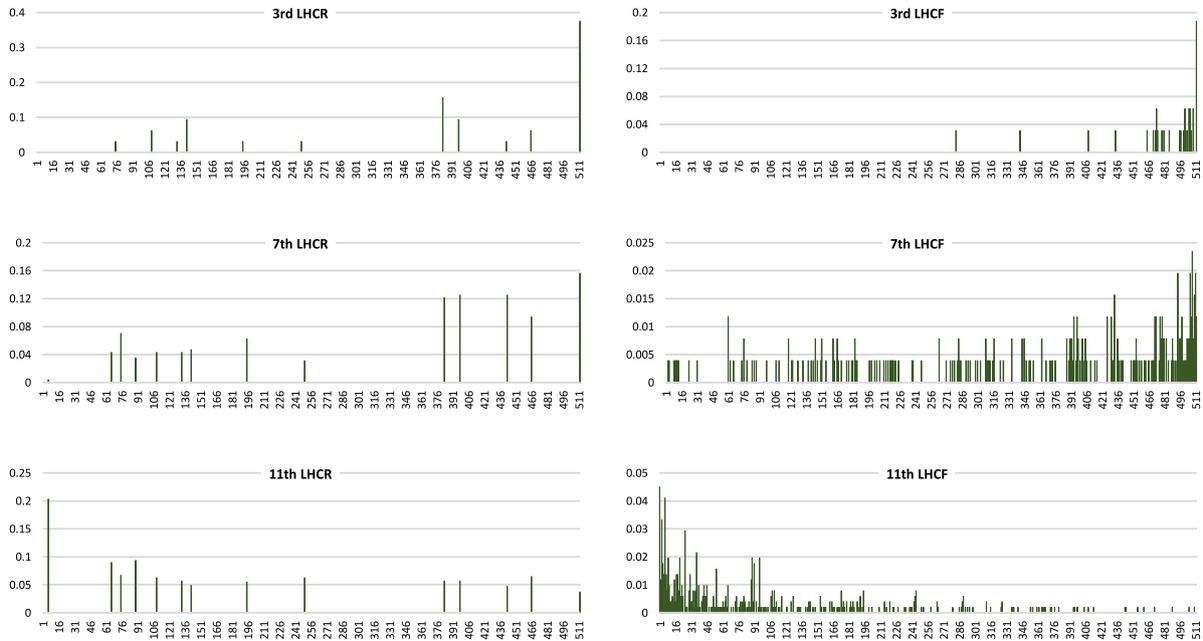

**Fig. 12.** Shape distribution of the 3rd/7th/11th layer of VGG16 built of LHCR and LHCF respectively. Left: LHCR; right: LHCF. The *x*-axis is the serial index of all 512 shapes; *y*-axis is the ratio a shape takes up in an LHC layer.

By comparing LHC with HetConv/FPGM/Taylor, the networks built of LHC always achieve better accuracy at any computation reduction ratio. Given the same computation reduction, ours accuracy surpasses others by 2.0% on CIFAR10 or 1.0% on ImageNet. For example, our method can improve VGG16 computation efficiency by nearly 5× on CIFAR10 or 2× on ImageNet without harming its performance.

Interestingly, there is always appropriate $\Delta$flops value, at which networks built of LHC surprisingly outperform the original, i.e., the horizontal dashed blue line in every sub-figure. This rarely happens with existing techniques. Specifically, our method improves top1 classification accuracy by 1.0% on CIFAR10 or 0.5% on ImageNet at most. Thus, our method is also a powerful training technique.

The parameter reduction is directly up to the $d_t$ we set, namely, $d_t \in \{$invalid, 0.2, 0.1, 0.05, 0.01$\}$ for CIFAR10, and $d_t \in \{$invalid, 0.25, 0.1$\}$ for ImageNet. Without harming the accuracy, our method can compress VGG/ResNet weights by about 10× on CIFAR10, or 4× on ImageNet.

We also conducted the hardware acceleration simulation, mentioned in Section 3.4, on a VGG16 model, which is trained on CIFAR10 under $d_t$ = 0.1 and achieves $\Delta$flops = 78% and *acc* = 91.45% (almost the same as the original performance). The number of clocks and memory rows (each row contains $3*3*c_{gi}*c_{go}$ weights) required during the inference of an input image of shape 32*32 are drawn in Fig. 11. In hardware implementation, the number of clocks and memory rows can also be saved by 5× and 10×. With less requirements in clock and memory, the power dissipation of hardware is accordingly reduced.

**LHC vs GWC+Channel-Shuffle, DWC+PWC & Inception**

Here $\Delta$flops are evaluated when ShuffleNet/MobileNet/GoogLeNet and their LHC variants have similar accuracy.

The GoogLeNet model, with its Inceptions rebuilt by LHC, achieves computation reduction $\Delta$flops = 24.39%, and parameter compression $\Delta$param = 64.56%; the MobileNet model, with its DWC+PWC modules replaced, achieves $\Delta$flops = 15.68% and $\Delta$params = 18.82%; and the ShuffleNet model, with its GWC+channel-shuffle modules replaced, achieves $\Delta$flops = 5.36% and $\Delta$params = 11.73%.

Clearly, the efficiency of GoogLeNet/ShuffleNet/MobileNet, either well-designed or light-weight, can be further improved by LHC, which means the efficiency gain contributed by our layer-level method even beats that of beyond-layer-level methods.

Interestingly, the computation reduction amplitude of our method on ShuffleNet is obviously less than on the other two, which verifies the necessity of information exchange provided by channel-shuffle.

## 6. Discussions

Here we discuss why LHC works. The VGG16 models trained on ImageNet with $d_t$ = invalid are chosen, since other models have similar results.

**What LHC Learns**

The core difference between LHC and standard convolution is the dendritic topology plasticity, i.e., learning masks of suitable shapes for current task in a data-driven manner. We count the number of shapes learnt by the 3rd/7th/11th LHC layer, as shown in Fig. 12. The *x*-axis is the serial indexes of all 512 shapes drawn in Fig. 3, and the *y*-axis is the ratio a shape takes up in an LHC layer.

By comparing the shape distribution of different LHCR/LHCF layers, we find that: (1) low layers prefer denser shapes and mainly have no shape ⟨1⟩1, which is totally sparse; (2) high layers have the least amount of denser shapes but the most amount of shape ⟨1⟩1; (3) middle layers takes the intermediate shape distribution.

This means that enough number of kernels in low layers is necessary for extracting sufficient basic patterns, yet after the non-linearity of a layer upon a layer, the patterns in the feature maps become increasingly abstract and sparse, and thus just need sparser kernels to extract. **How LHC Converges**

We penetrate into the training process by inspecting how the shapes/masks in LHC layers evolve along with the training epochs. We calculate the masks correlation like Guillaume et al. (2018) of an LHC layer between two adjacent epochs, i.e., the mean of element-wise-logical-and. We choose the 3rd/7th/11th layers of VGG16 built of LHCF to visualize in Fig. 13; LHCR has similar results. The *x*-axis is the number of epochs and the *y*-axis is the correlation of masks between two epochs.

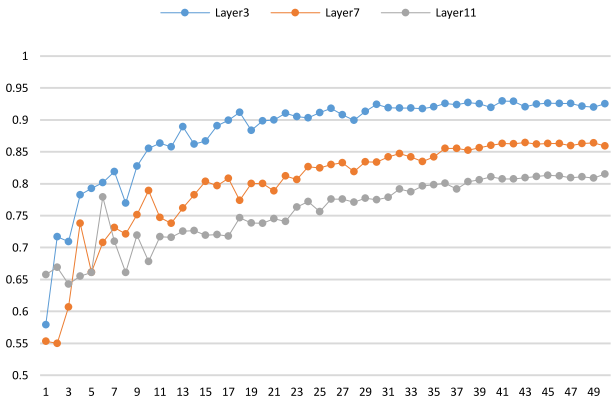

**Fig. 13.** Evolution of masks along with the training epochs of the 3rd/7th/11th LHCF layer. The *x*-axis is the epoch number; *y*-axis is the masks correlation. Mask correlation grows as training goes, which means LHC masks generally evolve from dynamic states to more static states.

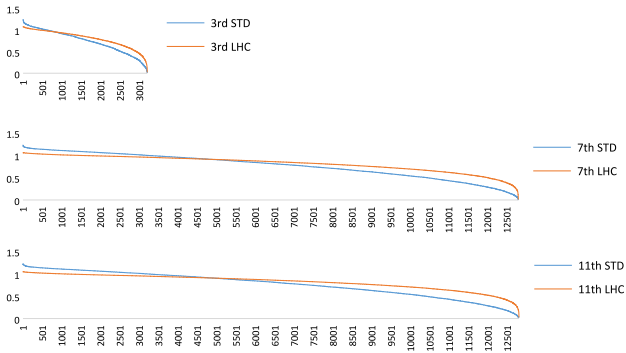

**Fig. 14.** The spectrums of the 3rd/7th/11th LHC layers and corresponding standard convolution layers from two models that are also used in Fig. 11. The *x*-axis is spectrum components, and *y*-axis is their amplitudes. The more the convolution layers' spectrum of a CNN model approximates uniform distribution, the better the model's performance will be. Apparently, LHC layers' spectrums are always more uniformly distributed than that of standard convolutions, which intuitively explains why under appropriate conditions models based on our method have performance better than models based on the standard convolution.

The correlation of masks in each layer increases as the training epochs carry forward, but is always less than 100%, even if the model converges enough. This means our differentiable step function keeps the dendritic topology updating all the time (of course so is the synaptic strength), just like the highly dynamic biological neural networks in the brain (Holtmaat et al., 2005; Stettler et al., 2006).

Besides, the correlation of lower layers is larger than the higher, which is also as expected, because the lower layers learn the patterns that are more task-agnostic and thus requires more stable dendritic topology.

**Why LHC Excels**

As pointed out by OCNN (Wang, Chen, Chakraborty, & Yu, 2020), the spectrum of a convolution layer's DBT (doubly block Toeplitz) matrix reasonably indicates how well the CNN model's capacity is utilized. The more uniformly the spectrums distribute, the better the model performances. Clearly drawn in Fig. 14, our method's spectrums are always closer to uniform distribution than the original. Combined with Fig. 13, we may assure that it is the changing masks that force the model to utilize its capacity better.

Now we know about the weights in a model built of LHC:

(1) there are a lot of zeros, up to 80+%, which can be taken as a case of L0/L1 regularization training, i.e., strength regularization;

(2) these zeros distribute in a structural way in the kernels, which further provides another kind of regularization, i.e., topology regularization;

(3) the position of these zeros changes along with the training steps, which can be seen as the dropout of the weights.

Hence, it is not difficult to understand why our method surpasses existing methods.

## 7. Conclusion

Our method LHC integrates the plasticity of both dendritic topology and synaptic strength, making both the kernel shapes and the weights learnable during training. CNN models built of LHC can be greatly sparsified structurally and can be accelerated in high parallelism. Experiments against existing convolution techniques show that LHC always achieves better performance at any computation reduction ratio; and experiments of rebuilding typical network structures show that LHC can improve their efficiency even further. Our method also achieves better CNN performance if taken as a training technique.

## CRediT authorship contribution statement

**Rongzhen Zhao:** Conceptualization, Methodology, Software. **Zhenzhi Wu:** Methodology. **Qikun Zhang:** Software.

## Declaration of competing interest

The authors declare that they have no known competing financial interests or personal relationships that could have appeared to influence the work reported in this paper.

## Acknowledgments

This work was supported by "Science and Technology Innovation 2030 - New Generation of Artificial Intelligence", China project (2020AAA0109100) and Beijing Science and Technology Plan, China (Z191100007519009).

## References

Alex, K., Sutskever, I., & Hinton, G. E. (2012). Imagenet classification with deep convolutional neural networks. In *Advances in neural information processing systems 25*.

Andri, R., Karunaratne, G., Cavigelli, L., & Benini, L. (2020). ChewBaccaNN: A flexible 223 TOPS/W BNN accelerator. arXiv preprint arXiv:2005.07137.

Beysolow II, T. (2017). Convolutional neural networks (CNNs/ConvNets). https://cs231n.github.io/convolutional-networks/.

Bhatt, D. H., Zhang, S., & Gan, W. (2009). Dendritic spine dynamics. *Annual Review of Physiology*, 71(1), 261–282.

Cao, S., Ma, L., Xiao, W., Zhang, C., Liu, Y., Zhang, L., et al. (2019). SeerNet: Predicting convolutional neural network feature-map sparsity through low-bit quantization. In *IEEE conference on computer vision and pattern recognition*.

Chen, X., Xie, L., Wu, J., & Tian, Q. (2019). Progressive differentiable architecture search: Bridging the depth gap between search and evaluation. In *IEEE international conference on computer vision*.

Christos, L., Max, W., & Diederik, P. K. (2018). Learning sparse neural networks through $L_0$ regularization. In *International conference on learning representations*.

Deng, B. L., Li, G., Han, S., Shi, L., & Xie, Y. (2020). Model compression and hardware acceleration for neural networks: A comprehensive survey. *Proceedings of the IEEE*, 108(4), 485–532.

Gong, R., Liu, X., Jiang, S., Li, T., Hu, P., Lin, J., et al. (2019). Differentiable soft quantization: Bridging full-precision and low-bit neural networks. In *IEEE international conference on computer vision*.

Goodfellow, I., Bengio, Y., & Courville, A. (2016). *Deep learning*. MIT press.

Guillaume, B., David, K., Wolfgang, M., & Robert, L. (2018). Deep rewiring: Training very sparse deep networks. In *International conference on learning representations*.

Harms, K. J., & Dunaevsky, A. (2007). Dendritic spine plasticity: Looking beyond development. *Brain Research*, 1184(1), 65–71.

He, Y., Liu, P., Wang, Z., Hu, Z., & Yang, Y. (2019). Filter pruning via geometric median for deep convolutional neural networks acceleration. In *IEEE conference on computer vision and pattern recognition* (pp. 4340–4349).

He, K., Zhang, X., Ren, S., & Sun, J. (2016). Deep residual learning for image recognition. In *IEEE conference on computer vision and pattern recognition*.

Holtmaat, A., Trachtenberg, J. T., Wilbrecht, L., Shepherd, G. M., Zhang, X., Knott, G., et al. (2005). Transient and persistent dendritic spines in the neocortex in vivo. *Neuron*, 45(2), 279–291.

Howard, A., Sandler, M., Chu, G., Chen, L., Chen, B., Tan, M., et al. (2019). Searching for MobileNetV3. In *IEEE international conference on computer vision*.

Howard, A. G., Zhu, M., Chen, B., Kalenichenko, D., Wang, W., Weyand, T., et al. (2017). Mobilenets: Efficient convolutional neural networks for mobile vision applications. arXiv preprint arXiv:1704.04861.

Hu, J., Shen, L., & Sun, G. (2018). Squeeze-and-excitation networks. In *IEEE conference on computer vision and pattern recognition*.

Huang, G., Liu, S., M.L.V., Der, & Weinberger, K. Q. (2018). CondenseNet: An efficient denseNet using learned group convolutions. In *IEEE conference on computer vision and pattern recognition*.

Jin, X., Peng, B., Wu, Y., Liu, Y., Liu, J., Liang, D., et al. (2019). Knowledge distillation via route constrained optimization. In *IEEE international conference on computer vision*.

Liu, Z., Mu, H., Zhang, X., Guo, Z., Yang, X., Cheng, K., et al. (2019). MetaPruning: Meta learning for automatic neural network channel pruning. In *IEEE international conference on computer vision*.

Ma, N., Zhang, X., Zheng, H., & Sun, J. (2018). ShuffleNet V2: Practical guidelines for efficient CNN architecture design. In *European conference on computer vision*.

Merolla, P. A., Arthur, J. V., Alvarezicaza, R., Cassidy, A. S., Sawada, J., Akopyan, F., et al. (2014). A million spiking-neuron integrated circuit with a scalable communication network and interface. *Science*, 345(6197), 668–673.

Molchanov, P., Mallya, A., Tyree, S., Frosio, I., & Kautz, J. (2019). Importance estimation for neural network pruning. In *IEEE conference on computer vision and pattern recognition* (pp. 11264–11272).

Moons, B., Bankman, D., Yang, L., Murmann, B., & Verhelst, M. (2018). BinarEye: An always-on energy-accuracy-scalable binary CNN processor with all memory on chip in 28 nm CMOS. In *IEEE custom integrated circuits conference* (pp. 1–4).

Peng, Z., Li, Z., Zhang, J., Li, Y., Qi, G., & Tang, J. (2019). Few-Shot image recognition with knowledge transfer. In *IEEE international conference on computer vision*.

Peng, B., Tan, W., Li, Z., Zhang, S., Xie, D., & Pu, S. (2018). Extreme network compression via filter group approximation. In *European conference on computer vision* (pp. 300–316).

Sandler, M., Howard, A., Zhu, M., Zhmoginov, A., & Chen, L. (2018). MobileNetV2: Inverted residuals and linear bottlenecks. In *IEEE conference on computer vision and pattern recognition*.

Shang, W., Sohn, K., Almeida, D., & Lee, H. (2016). Understanding and improving convolutional neural networks via concatenated rectified linear units. In *International conference on machine learning* (pp. 2217–2225).

SIfre, L., & Mallat, S. (2014). *Rigid-Motion scattering for texture classification*. Pennsylvania State University.

Simonyan, K., & Zisserman, A. (2014). Very deep convolutional networks for large-scale image recognition. arXiv preprint arXiv:1409.1556.

Singh, P., Verma, V. K., Rai, P., & Namboodiri, V. P. (2019). HetConv: Heterogeneous kernel-based convolutions for deep CNNs. In *IEEE conference on computer vision and pattern recognition*.

Stettler, D. D., Yamahachi, H., Li, W., Denk, W., & Gilbert, C. D. (2006). Axons and synaptic boutons are highly dynamic in adult visual cortex. *Neuron*, 49(6), 877–887.

Sun, K., Li, M., Liu, D., & Wang, J. (2018). IGCV3: Interleaved low-rank group convolutions for efficient deep neural networks. In *British machine vision conference*.

Szegedy, C., Ioffe, S., Vanhoucke, V., & Alemi, A. A. (2016). *Inception-v4, Inception-ResNet and the impact of residual connections on learning*.

Szegedy, C., Liu, W., Jia, Y., Sermanet, P., Reed, S., Anguelov, D., et al. (2015). Going deeper with convolutions. In *IEEE conference on computer vision and pattern recognition*.

Szegedy, C., Vanhoucke, V., Ioffe, S., Shlens, J., & Wojna, Z. (2016). Rethinking the inception architecture for computer vision. In *IEEE conference on computer vision and pattern recognition*.

Tan, M., & Le, Q. V. (2019). Mixconv: Mixed depthwise convolutional kernels. arXiv preprint arXiv:1907.09595.

Verelst, T., & Tuytelaars, T. (2020). Dynamic convolutions: Exploiting spatial sparsity for faster inference. In *Proceedings of the IEEE/CVF conference on computer vision and pattern recognition* (pp. 2320–2329).

Wang, J., Chen, Y., Chakraborty, R., & Yu, X. (2020). Orthogonal convolutional neural networks. In *IEEE conference on computer vision and pattern recognition* (pp. 11505–11515).

Wang, Y., Xu, C., Chunjing, X., Xu, C., & Tao, D. (2018). Learning versatile filters for efficient convolutional neural networks. In *Advances in neural information processing systems* (pp. 1608–1618).

Xie, G., Wang, J., Zhang, T., Lai, J., Hong, R., & Qi, G. (2018). Interleaved structured sparse convolutional neural networks. In *IEEE conference on computer vision and pattern recognition*.

Yan, Z., Li, X., Li, M., Zuo, W., & Shan, S. (2018). Shift-net: Image inpainting via deep feature rearrangement. In *European conference on computer vision*.

Yin, H., Gong, Y., & Qiu, G. (2019). Side window filtering. In *IEEE conference on computer vision and pattern recognition*.

Zhang, J. (2019). Seesaw-Net: Convolution neural network with uneven group convolution. arXiv preprint arXiv:1905.03672.

Zhang, T., Qi, G., Xiao, B., & Wang, J. (2017). Interleaved group convolutions. In *IEEE international conference on computer vision*.

Zhang, X., Zhou, X., Lin, M., & Sun, J. (2018). ShuffleNet: An extremely efficient convolutional neural network for mobile devices. In *IEEE conference on computer vision and pattern recognition*.

Zhou, Y., Zhang, Y., Wang, Y., & Tian, Q. (2019). Accelerate CNN via recursive Bayesian pruning. In *IEEE international conference on computer vision*.